\documentclass[11pt]{article}

\usepackage[final]{acl}
\usepackage{authblk}
\usepackage{times}
\usepackage{latexsym}
\usepackage{booktabs}
\usepackage{amsmath}
\usepackage[T1]{fontenc}

\usepackage[utf8]{inputenc}

\usepackage{microtype}

\usepackage{inconsolata}
\usepackage{pifont}
\usepackage{graphicx}

\title{Be Faithful When Response: Returning Fluent and Grounded Answers for Vision-Language Models Reinforcement Learning}

\author[1]{$\text{Peng Gao}^{1,\dagger}$, $\text{Yujian Lee}^{2,\dagger}$, $\text{Yin Zhang}^3$, $\text{Yanglin Zhang}^3$, $\text{Haonan Wu}^3$, $\text{Zishan Liu}^3$,\\ $\text{Ruoxi Zang}^3$, $\text{Xin Zhu}^3$, $\text{Jiayin Zheng}^3$, $\text{Jian Yao}^3$, $\text{Zefeng Ji}^{3,*}$, $\text{Fei Ma}^1$ \\
\\
{$^{1}$GMLab,\quad $^{2}$Hong Kong Baptist University,\quad $^{3}$XPENG Robotics,}\\{$^{*}$Project leader,\quad $^{\dagger}$Equal Contribution}}

\begin{document}
\maketitle
\begin{abstract}
Reinforcement Learning (RL) is an important paradigm for improving the reasoning capabilities of Vision-Language Models (VLMs). However, directly applying RL to rollout multimodal reasoning can lead to instability, due to the exploitation of language priors, the neglect of visual evidence, and the generation of reasoning traces that are fluent yet not visually grounded. The question arises: Can initially steer the policy toward visually faithful reasoning regime before applying reinforcement learning? To this end, we propose a \textbf{Faithful Warm-Start (FWS)} strategy that first curates samples with explicit vision-language causal relationships from six general VQA benchmarks to construct the \textbf{FaithfulQA} dataset, where each of the image-question pairs gains a certain degree of visual observations, question requirements, commonsense knowledge, domain knowledge, and the final answer. Subsequently, a VLM-based judge is employed to further purify the dataset, ensuring strong causal consistency and visual faithfulness. This warm-start stage equips the model with the capability to understand causally grounded vision-language patterns before subsequent RL optimization under sparse answer-level rewards. Experimental results show that such faithful supervision improves answer accuracy, stabilizes RL training, and reduces visually unsupported reasoning.



\end{abstract}

\section{Introduction}
Vision-Language Models (VLMs) have demonstrated strong performance across a wide range of multimodal tasks, from visual question answering to complex visual reasoning \cite{lee2025optical, zhang2024vision, chen2025optimizing, gao2025contextual}. Recent advances further suggest that reinforcement learning (RL) is a promising paradigm for directly optimizing task-level rewards, thereby enhancing reasoning capabilities in a data-driven manner \cite{liu2025visual}. However, applying RL in the vision-language domain remains challenging, as reward signals are typically sparse and only evaluate final outcomes \cite{wu2025reinforcement}. As a result, models obtains high rewards by exploiting spurious cues rather than following faithful and visually grounded reasoning processes \cite{pan2026reward, wang2025comprehensive}.

This issue is particularly severe in vision-language reasoning tasks, where models often rely on dataset biases, language priors, or shortcut correlations, producing reasoning traces that function more as post-hoc rationalizations than genuine step-by-step derivations \cite{moll2025evaluating, kargupta2025cognitive}. During RL training, models can generate increasingly confident or verbose explanations without corresponding improvements in visual grounding. This ultimately leads to a fundamental misalignment between answer-level optimization and faithful multimodal reasoning, where the optimization objective prioritizes reward maximization over true reasoning consistency and visual adherence \cite{luo2026thinking, zheng2025deepeyes, gao2026remember}.
\begin{figure*}[t]
    \centering
    \includegraphics[width=0.92\textwidth]{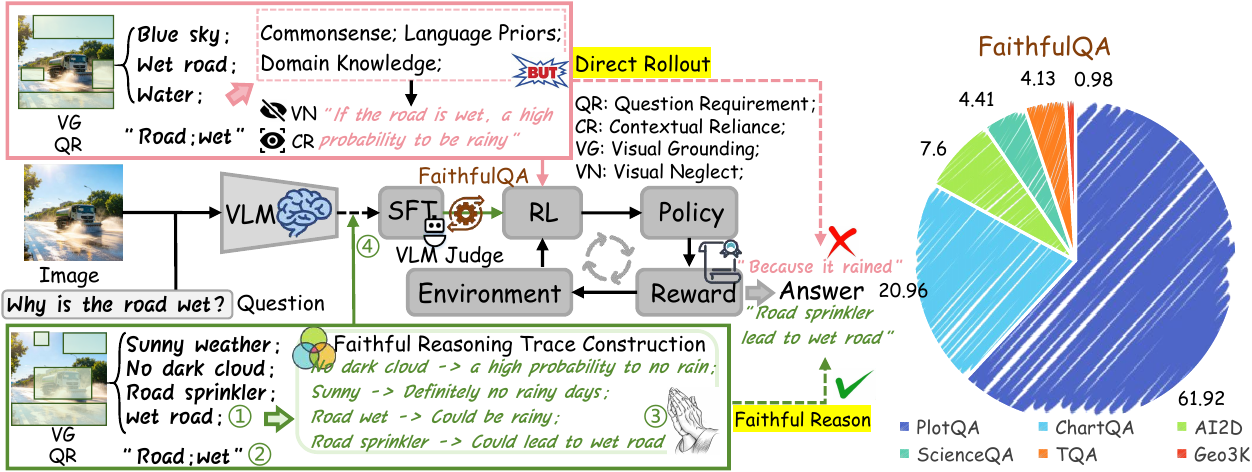}
    \caption{Pipeline of the proposed Faithful Workflow Strategy (FWS). Direct rollout typically suffers from a lack of visual grounding, leading to highly confident but erroneous outputs. FWS addresses this by eliciting faithful reasoning traces. Recognizing that standard VQA triplets often fail to convey true causal effects between vision and language, FWS screens six benchmarks to filter out data with weak causal links, composing the FaithfulQA dataset (\ding{172}$\sim$\ding{175}). After four stages of faithful trace elicitation, a advanced VLM judge evaluates and refines these traces during SFT based on their causal dependence on visual evidence. The resulting high-confidence, faithfully grounded dataset prevents the VLM from drifting into degenerate reasoning during subsequent RL training.}
    \label{fig:example}
\end{figure*}

In this work, we argue that the initial policy state is a pre-condition~\cite{li2025warmup} for successful vision-language RL~\cite{akarajaradwong2025cold,wei2025advancing}, yet often being overlooked. Directly initiating RL from a vanilla instruction-tuned VLM may place the policy in a suboptimal region of the optimization landscape, characterized by weak visual grounding and unstable reasoning trajectories. Starting from such a state, sparse rewards can further exacerbate shortcut learning, rather than encouraging genuine reasoning. To address this issue, we introduce a \textbf{Faithful Warm-Start (FWS)} strategy grounded in faithful visual reasoning traces. Specifically, as illustrated in Fig. \ref{fig:example}, FWS first constructs a raw \textbf{FaithfulQA} dataset from six general VQA benchmarks through a four-stage reasoning pipeline: \ding{172} Visual Grounding, \ding{173} Question Requirement, \ding{174} Intermediate Inference, and \ding{175} Final Prediction, enabling the model to learn causally grounded vision-language reasoning patterns. Subsequently, a VLM-based judge further refines these reasoning traces through rigorous filtering for visual faithfulness and logical consistency, thereby establishing a faithful initialization prior to RL optimization. Through extensive experiments on visual question answering and multimodal reasoning benchmarks, FWS improves answer accuracy, stabilizes RL training, and produces reasoning traces that are more faithfully grounded in visual evidence. These results suggest that faithful warm-start initialization is not merely a formatting choice, but a crucial prerequisite for effective reinforcement learning in VLMs. Our contributions are summarized as follows:
\begin{itemize}
    \item Highlighting the importance of policy initialization in vision-language RL and show that poor initialization promotes shortcut learning under sparse rewards.
    \item The proposed FWS strategy compose a FaithfulQA dataset, having it as a data-driven manner to have the model understand the causual patterns prior to RL optimization. 
    \item Empirical results show that FWS improves accuracy, training stability, and visual faithfulness across multimodal reasoning tasks.
\end{itemize}

\section{Related Work}
VLMs have demonstrated strong capabilities in complex multimodal reasoning tasks \cite{zhu2025internvl3, lee2025boosting, gao2025contextual, chen2025optimizing, gao2025ncl}. Recent works further explore RL to enhance such reasoning abilities through sparse rewards based on final answer correctness~\cite{kancheti2026faithful,tamo2026evidencerl,liu2026craft, gao2026remember}. However, directly applying RL to rollout multimodal reasoning in Fig. \ref{fig:example} often leads to training instability and shortcut learning, where models rely on language priors or dataset biases rather than visual evidence \cite{zhang2026seeing,li2026palmr}. As a result, ensuring faithful reasoning remains a fundamental challenge \cite{cao2025reveal,uppaal2026journey}. Existing methods mainly focus on reward design during RL, while overlooking the importance of policy initialization. We argue that weakly grounded initial policies can drive optimization toward degenerate reasoning behaviors under sparse rewards. \textbf{Motivation}. We propose that a faithful supervised warm-start is crucial for establishing a visually grounded and stable reasoning regime before RL optimization begins.

\section{Methodology}
The FWS strategy illustrated in Fig. \ref{fig:example} aims to equip the a baseline VLM (Qwen3-VL-2B-Instruct) with faithful reasoning traces, preventing the model from drifting toward degenerate reasoning behaviors during RL optimization. To encourage the model to learn causally grounded vision-language patterns, FWS first curates samples from six general VQA benchmarks and constructs faithful reasoning traces through four stages of faithful trace construction (\ding{172}$\sim$\ding{175}). These traces are then rigorously verified for their causal dependence on visual evidence, and subsequently refined through SFT under the guidance of Qwen3VL-8B-Thinking \cite{bai2025qwen3}, having it to be a VLM-based judge, ultimately constructing a high-confidence dataset of faithful multimodal reasoning traces. We claim that FaithfulQA is used solely for warm-start stage, while the original benchmark datasets are strictly preserved for inference and evaluation.


\subsection{Faithful Trace Construction} \label{pre}
\paragraph{Rollout and pre-Filtering} The rollout stage externalizes the model’s latent reasoning behavior into explicit reasoning trajectories. Given an image, a multimodal question, and candidate options, the VLM is prompted to generate a comprehensive reasoning process within \texttt{<think>} tags before producing the final prediction within \texttt{<answer>} tags. To facilitate fine-grained causal analysis, each reasoning trajectory is further decomposed into a sequence of structured \ding{172}$\sim$\ding{175}\textit{``Reasoning Units''}. Each unit corresponds to an atomic reasoning operation and is explicitly categorized into one of four categories: \ding{172}\textit{``Visual Grounding''} (direct evidence observed from the image), \ding{173}\textit{``Question Requirement''} (external commonsense or domain knowledge), \ding{174}\textit{``Intermediate Inference''} (intermediate logical steps derived from facts and rules), and \ding{175}\textit{``Final Prediction''} (the final derivation supporting the chosen answer). Following the rollout phase, we perform an initial filtering stage based on answer correctness, retaining only trajectories that lead to correct predictions, ensuring subsequent causal analysis is focused on decomposing the logic behind successful decisions, effectively isolating ``right for the right reasons'' traces from spurious successes.

\subsection{Faith Consistency Examination}
\paragraph{Unit-level Causal Intervention} The core of our approach lies in verifying whether the generated reasoning units are functionally necessary for the model’s final decision, rather than being mere post-hoc narrations. For a correct reasoning trace consisting of $N$ units, we perform $N$ separate intervention trials. In each trial, we remove exactly one reasoning unit to form a \textit{``partial reasoning trace''}. Having it alongside with the original image and question, are re-fed into the VLM to obtain a new answer. By observing how the model's performance degrades when a specific piece of information is withheld, we can empirically quantify the causal weight of each unit. We define a primary metric, \texttt{num\_correct\_to\_wrong}, to capture the causal impact of the reasoning units. This metric represents the number of times the model's prediction transitions from correct to incorrect upon the removal of a reasoning unit. A non-zero value for this metric provides direct causal evidence that the reasoning units are not redundant descriptions but are actively grounding the final decision.

\subsection{Faithful Trace Selection}\label{post}
\paragraph{Judging and post-Filtering} To ensure high data quality, we employ the Qwen3VL-8B-Thinking as the judge to evaluate refined reasoning traces across four critical dimensions, each scored on a scale of $0$ to $2$. The evaluation includes: \texttt{visual\_grounding\_score}, measuring dependence on actual visual evidence; \texttt{unit\_causality\_score}, assessing the logical and causal necessity of reasoning units based on intervention results; \texttt{non\_contradiction\_score}, checking internal logical consistency and consistency with the image; and \texttt{faithfulness\_score}, which provides an overall assessment of whether the reasoning chain faithfully explains the decision process. The final faithful reasoning traces are obtained through strict post-filtering, requiring a perfect score of $2$ across all four dimensions and a causal metric strictly greater than zero. Finally, when multiple candidate rollouts are available for the same question, we perform deduplication by selecting the trace with the strongest causal evidence and highest alignment scores. This procedure ensures a diverse yet high-quality FaithfulQA dataset at warmup stage for RL optimization.

\section{Experiments}
\label{sec:experiments}
Extensive experiments are conducted to comprehensively evaluate the effectiveness of the FWS strategy. Specifically, our study focuses on the following research questions:
(1) Can FWS effectively improve the visual faithfulness and causal grounding of reasoning traces?
(2) How does FWS influence the stability and robustness of the subsequent RL training process?


\subsection{Experimental Setup}
\label{sec:setup}
\textbf{Datasets}. We construct FaithfulQA from six representative multimodal reasoning benchmarks, including AI2D~\cite{DBLP:conf/eccv/KembhaviSKSHF16}, TextbookQA~\cite{Kembhavi_2017_CVPR}, PlotQA~\cite{Methani_2020_WACV}, ScienceQA~\cite{lu2022learn}, ChartQA~\cite{masry-etal-2022-chartqa}, and Geo3K~\cite{lu-etal-2021-inter}, resulting in a total of 137,568 samples. These datasets cover diverse reasoning scenarios, such as chart comprehension, diagram reasoning, scientific question answering, and geometry problem solving. For each image–question pair, we first generate reasoning traces, yielding 121,866 samples after pre-filtering (\S\ref{pre}). We then perform answer-correctness filtering based on unit-level interventions using a VLM-based judge under the SFT policy, ultimately retaining 60,000 high-quality samples after post-filtering (\S\ref{post}). In addition, to evaluate whether the proposed warm-start strategy can improve subsequent reinforcement learning, we sample another 20,000 non-overlapping instances from the remaining data pool after SFT data selection and use them exclusively for the RL stage. Final model evaluation is still conducted on the original benchmark datasets. 

\textbf{Baselines}. We use Qwen3-VL-Instruct~\cite{bai2025qwen3} as backbone. Unless otherwise specified, all methods are initialized from the same pretrained checkpoint to ensure fair comparison. FWS is applied prior to RL optimization, while all methods share the same reward design and training settings during the RL stage.

\subsection{Main Results}
\label{sec:results}

\begin{table}[t]
\centering
\small
\caption{Comparison (\textit{w.} and \textit{w/o.} \textit{RL}) of different SFT data scales on five multimodal reasoning benchmarks.}
\label{tab:sft_scale}
\resizebox{0.485\textwidth}{!}{
\begin{tabular}{lcccccc}
\toprule
Model & AI2D & TQA & PlotQA & SciQA & ChartQA & Avg. \\
\midrule
\multicolumn{7}{c}{\textit{w/o. RL} }\\ 
\midrule
Base model & 60.33 & 69.49 & 56.72 & 78.38  & 89.64 & 70.91 \\
20K-SFT    & 68.94 & 74.09 & \textbf{70.77} & 86.56  & 94.64 & \textbf{79.00}\\
40K-SFT    & 68.71 & 74.80 & 68.41 & 86.26  & 94.68 & 78.57 \\
60K-SFT    & \textbf{69.43} & \textbf{75.01} & 67.76 & \textbf{87.65}  & \textbf{94.80} & 78.93 
\\
\midrule
\multicolumn{7}{c}{\textit{w. RL} }\\ 
\midrule
Base model          & 60.00 & 69.77 & 57.95 & 78.18 & 89.48 & 71.07 \\
20K-SFT         & 69.23 & 74.61 & \textbf{70.80} & 86.81 & 96.12 & \textbf{79.51} \\
40K-SFT         & 69.23 & \textbf{75.28} & 68.78 & 86.86 & 96.08 & 79.24 \\
60K-SFT        & \textbf{69.36} & \textbf{75.28} & 67.97 & \textbf{87.90} & \textbf{96.60} & 79.42 \\
\bottomrule
\end{tabular}}
\end{table}


Results in Table~\ref{tab:sft_scale} evaluate whether FWS improves both SFT and subsequent RL optimization under three faithful SFT data scales (20K, 40K, and 60K). FWS consistently improves the base model across all benchmarks, with even 20K samples yielding clear gains, demonstrating the effectiveness of faithful reasoning supervision prior to RL. Increasing the SFT scale further enhances performance, while 60K-SFT approaches saturation on most benchmarks. Moreover, RL training initialized from FWS-supervised checkpoints consistently outperforms the base RL model, highlighting the importance of faithful warm-start initialization. Rather than learning grounded reasoning solely from sparse rewards, FWS first steers the model into a more faithful reasoning regime, providing a stronger and more stable starting point for subsequent RL optimization. Overall, these results support the hypothesis that faithful supervised warm-start is critical for effective VLMs RL.

\begin{figure}[t]
    \centering
    \includegraphics[width=0.5\textwidth]{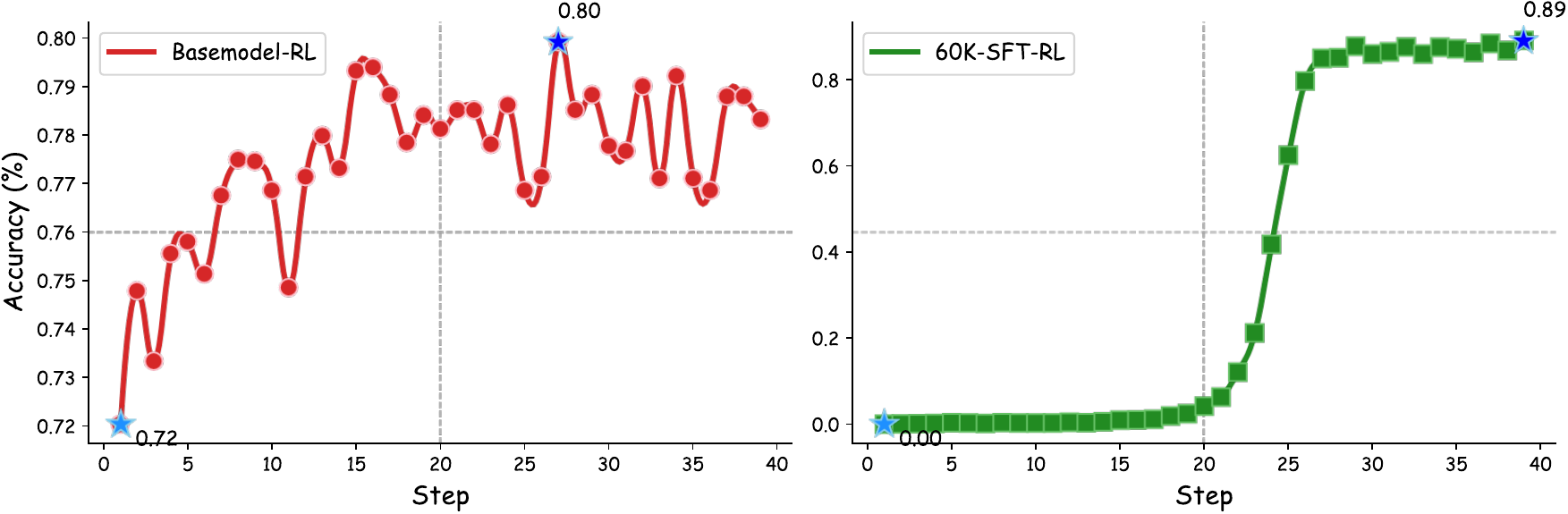}
    \caption{Reward dynamics of Basemodel-RL (A) and 60K-SFT-RL (B). }
    \label{curve}
\end{figure}

\subsection{Analysis of Reasoning Faithfulness}
\label{sec:faithfulness}
We further analyze the reward dynamics during RL training to understand the impact of faithful warm-start initialization. In Fig.~\ref{curve}A, RL directly initialized from the base model exhibits only gradual improvements, with rewards increasing in the early stage but eventually saturating at a relatively low plateau accompanied by noticeable fluctuations. In contrast, Fig.~\ref{curve}B shows RL initialized from the 60K faithful SFT checkpoint follows a markedly different trajectory, where rewards remain low during an initial adaptation phase, after which they rise sharply and rapidly converge to a higher and more stable plateau. This behavior suggests that faithful SFT encodes informative reasoning priors that can be effectively activated and refined by RL, enabling the policy to more efficiently discover reward-aligned reasoning trajectories and achieve superior optimization performance.


\section{Future Work}

In future work, we plan to further extend FWS by integrating the reasoning dynamics acquired during the SFT stage into the RL reward modeling process. Instead of relying exclusively on outcome-level supervision based on final answer correctness, we aim to incorporate process-level rewards that explicitly encourage visually grounded, causally coherent, and logically consistent intermediate reasoning trajectories. Such a design has the potential to further enhance the stability and faithfulness of multimodal reinforcement learning.

\newpage
\bibliography{custom}

@inproceedings{lee2025boosting,
  title={Boosting Audio-Visual Segmentation via Triple-Modalities Alignment},
  author={Lee, Yujian and Gao, Peng and Chen, Zailong and Fan, Wentao and Jing, Guquan and Hu, Yiyang},
  booktitle={2025 IEEE International Conference on Multimedia and Expo (ICME)},
  pages={1--6},
  year={2025},
  organization={IEEE}
}

@inproceedings{lee2025optical,
  title={How Do Optical Flow and Textual Prompts Collaborate to Assist in Audio-Visual Semantic Segmentation?},
  author={Lee, Yujian and Gao, Peng and Xu, Yongqi and Fan, Wentao},
  booktitle={Proceedings of the IEEE/CVF International Conference on Computer Vision},
  pages={23342--23352},
  year={2025}
}

@article{zhang2024vision,
  title={Vision-language models for vision tasks: A survey},
  author={Zhang, Jingyi and Huang, Jiaxing and Jin, Sheng and Lu, Shijian},
  journal={IEEE transactions on pattern analysis and machine intelligence},
  volume={46},
  number={8},
  pages={5625--5644},
  year={2024},
  publisher={IEEE}
}

@inproceedings{liu2025visual,
  title={Visual-rft: Visual reinforcement fine-tuning},
  author={Liu, Ziyu and Sun, Zeyi and Zang, Yuhang and Dong, Xiaoyi and Cao, Yuhang and Duan, Haodong and Lin, Dahua and Wang, Jiaqi},
  booktitle={Proceedings of the IEEE/CVF International Conference on Computer Vision},
  pages={2034--2044},
  year={2025}
}

@article{wu2025reinforcement,
  title={Reinforcement Learning for Large Model: A Survey},
  author={Wu, Weijia and Gao, Chen and Chen, Joya and Lin, Kevin Qinghong and Meng, Qingwei and Zhang, Yiming and Qiu, Yuke and Zhou, Hong and Shou, Mike Zheng},
  journal={arXiv preprint arXiv:2508.08189},
  year={2025}
}

@article{pan2026reward,
  title={Reward modeling for reinforcement learning-based LLM reasoning: Design, challenges, and evaluation},
  author={Pan, Pei-Chi and Liang, Yingbin and Lin, Sen},
  journal={arXiv preprint arXiv:2602.09305},
  year={2026}
}

@article{wang2025comprehensive,
  title={A comprehensive survey on trustworthiness in reasoning with large language models},
  author={Wang, Yanbo and Yu, Yongcan and Liang, Jian and He, Ran},
  journal={arXiv preprint arXiv:2509.03871},
  year={2025}
}

@article{moll2025evaluating,
  title={Evaluating Reasoning Faithfulness in Medical Vision-Language Models using Multimodal Perturbations},
  author={Moll, Johannes and Graf, Markus and Lemke, Tristan and Lenhart, Nicolas and Truhn, Daniel and Delbrouck, Jean-Benoit and Pan, Jiazhen and Rueckert, Daniel and Adams, Lisa C and Bressem, Keno K},
  journal={arXiv preprint arXiv:2510.11196},
  year={2025}
}

@article{kargupta2025cognitive,
  title={Cognitive foundations for reasoning and their manifestation in llms},
  author={Kargupta, Priyanka and Li, Shuyue Stella and Wang, Haocheng and Lee, Jinu and Chen, Shan and Ahia, Orevaoghene and Light, Dean and Griffiths, Thomas L and Kleiman-Weiner, Max and Han, Jiawei and others},
  journal={arXiv preprint arXiv:2511.16660},
  year={2025}
}

@article{luo2026thinking,
  title={When thinking drifts: Evidential grounding for robust video reasoning},
  author={Luo, Romy and Xue, Zihui Sherry and Dimakis, Alex and Grauman, Kristen},
  journal={Advances in Neural Information Processing Systems},
  volume={38},
  pages={83696--83727},
  year={2026}
}

@article{zheng2025deepeyes,
  title={Deepeyes: Incentivizing" thinking with images" via reinforcement learning},
  author={Zheng, Ziwei and Yang, Michael and Hong, Jack and Zhao, Chenxiao and Xu, Guohai and Yang, Le and Shen, Chao and Yu, Xing},
  journal={arXiv preprint arXiv:2505.14362},
  year={2025}
}

@article{bai2025qwen3,
  title={Qwen3-vl technical report},
  author={Bai, Shuai and Cai, Yuxuan and Chen, Ruizhe and Chen, Keqin and Chen, Xionghui and Cheng, Zesen and Deng, Lianghao and Ding, Wei and Gao, Chang and Ge, Chunjiang and others},
  journal={arXiv preprint arXiv:2511.21631},
  year={2025}
}

@InProceedings{Methani_2020_WACV,
  author = {Methani, Nitesh and Ganguly, Pritha and Khapra, Mitesh M. and Kumar, Pratyush},
  title = {PlotQA: Reasoning over Scientific Plots},
  booktitle = {Proceedings of the IEEE/CVF Winter Conference on Applications of Computer Vision (WACV)},
  month = {March},
  year = {2020}
}

@inproceedings{masry-etal-2022-chartqa,
  title = "{C}hart{QA}: A Benchmark for Question Answering about Charts with Visual and Logical Reasoning",
  author = "Masry, Ahmed and Long, Do Xuan and Tan, Jia Qing and Joty, Shafiq and Hoque, Enamul",
  booktitle = "Findings of the Association for Computational Linguistics: ACL 2022",
  month = may,
  year = "2022",
  address = "Dublin, Ireland",
  publisher = "Association for Computational Linguistics",
  url = "https://aclanthology.org/2022.findings-acl.177/",
  doi = "10.18653/v1/2022.findings-acl.177",
  pages = "2263--2279"
}

@inproceedings{DBLP:conf/eccv/KembhaviSKSHF16,
  author = {Aniruddha Kembhavi and Mike Salvato and Eric Kolve and Min Joon Seo and Hannaneh Hajishirzi and Ali Farhadi},
  title = {A Diagram is Worth a Dozen Images},
  booktitle = {Computer Vision - ECCV 2016 - 14th European Conference, Amsterdam, The Netherlands, October 11-14, 2016, Proceedings, Part IV},
  series = {Lecture Notes in Computer Science},
  pages = {235--251},
  publisher = {Springer},
  year = {2016},
  doi = {10.1007/978-3-319-46493-0_15}
}

@inproceedings{lu2022learn,
  title = {Learn to Explain: Multimodal Reasoning via Thought Chains for Science Question Answering},
  author = {Lu, Pan and Mishra, Swaroop and Xia, Tony and Qiu, Liang and Chang, Kai-Wei and Zhu, Song-Chun and Tafjord, Oyvind and Clark, Peter and Kalyan, Ashwin},
  booktitle = {The 36th Conference on Neural Information Processing Systems (NeurIPS)},
  year = {2022}
}

@InProceedings{Kembhavi_2017_CVPR,
  author = {Kembhavi, Aniruddha and Seo, Minjoon and Schwenk, Dustin and Choi, Jonghyun and Farhadi, Ali and Hajishirzi, Hannaneh},
  title = {Are You Smarter Than a Sixth Grader? Textbook Question Answering for Multimodal Machine Comprehension},
  booktitle = {Proceedings of the IEEE Conference on Computer Vision and Pattern Recognition (CVPR)},
  month = {July},
  year = {2017}
}

@inproceedings{lu-etal-2021-inter,
  title = "{I}nter-{GPS}: Interpretable Geometry Problem Solving with Formal Language and Symbolic Reasoning",
  author = "Lu, Pan and Gong, Ran and Jiang, Shibiao and Qiu, Liang and Huang, Siyuan and Liang, Xiaodan and Zhu, Song-Chun",
  booktitle = "Proceedings of the 59th Annual Meeting of the Association for Computational Linguistics and the 11th International Joint Conference on Natural Language Processing (Volume 1: Long Papers)",
  month = aug,
  year = "2021",
  address = "Online",
  publisher = "Association for Computational Linguistics",
  url = "https://aclanthology.org/2021.acl-long.528/",
  doi = "10.18653/v1/2021.acl-long.528",
  pages = "6774--6786"
}

@article{zhu2025internvl3,
  title={Internvl3: Exploring advanced training and test-time recipes for open-source multimodal models},
  author={Zhu, Jinguo and Wang, Weiyun and Chen, Zhe and Liu, Zhaoyang and Ye, Shenglong and Gu, Lixin and Tian, Hao and Duan, Yuchen and Su, Weijie and Shao, Jie and others},
  journal={arXiv preprint arXiv:2504.10479},
  year={2025}
}

@article{kancheti2026faithful,
  title={Faithful GRPO: Improving Visual Spatial Reasoning in Multimodal Language Models via Constrained Policy Optimization},
  author={Kancheti, Sai Srinivas and Kanade, Aditya and Sinha, Rohit and Balasubramanian, Vineeth N and Ganu, Tanuja},
  journal={arXiv preprint arXiv:2604.08476},
  year={2026}
}

@article{tamo2026evidencerl,
  title={EvidenceRL: Reinforcing Evidence Consistency for Trustworthy Language Models},
  author={Tamo, J Ben and Lu, Yuxing and Marteau, Benoit L and Nnamdi, Micky C and Wang, May D},
  journal={arXiv preprint arXiv:2603.19532},
  year={2026}
}

@article{liu2026craft,
  title={CRAFT: Calibrated Reasoning with Answer-Faithful Traces via Reinforcement Learning for Multi-Hop Question Answering},
  author={Liu, Yu and Zhang, Wenxiao and Cao, Cong and Yuan, Fangfang and Chen, Weizhuo and Hu, Cheng and Xu, Pin and Yang, Yuling and Peng, Kun and Guo, Diandian and others},
  journal={arXiv preprint arXiv:2602.01348},
  year={2026}
}

@article{zhang2026seeing,
  title={Seeing is Believing? Mitigating OCR Hallucinations in Multimodal Large Language Models},
  author={Zhang, Can and Wu, Ziheng and Chen, Zhenghao and Zhan, Yufei and Li, Yifan and Zhang, Zhao and Wang, Xian and Qiu, Minghui and others},
  journal={Advances in Neural Information Processing Systems},
  volume={38},
  pages={74230--74248},
  year={2026}
}

@article{li2026palmr,
  title={PaLMR: Towards Faithful Visual Reasoning via Multimodal Process Alignment},
  author={Li, Yantao and Hui, Qiang and Yan, Chenyang and Cheng, Kanzhi and Zhao, Fang and Tan, Chao and Gao, Huanling and Zhang, Jianbing and Wang, Kai and Dai, Xinyu and others},
  journal={arXiv preprint arXiv:2603.06652},
  year={2026}
}

@article{cao2025reveal,
  title={REVEAL: Reasoning-Enhanced Forensic Evidence Analysis for Explainable AI-Generated Image Detection},
  author={Cao, Huangsen and Mei, Qin and Li, Zhiheng and Li, Yuxi and Meng, Zhan and Zhang, Ying and Li, Chen and Zhang, Zhimeng and Ding, Xin and Wang, Yongwei and others},
  journal={arXiv preprint arXiv:2511.23158},
  year={2025}
}

@inproceedings{uppaal2026journey,
  title={Journey Before Destination: On the importance of Visual Faithfulness in Slow Thinking},
  author={Uppaal, Rheeya and Htut, Phu Mon and Bai, Min and Pappas, Nikolaos and Qi, Zheng and Swamy, Sandesh},
  booktitle={Proceedings of the 19th Conference of the European Chapter of the Association for Computational Linguistics (Volume 1: Long Papers)},
  pages={4147--4168},
  year={2026}
}

@inproceedings{li2025warmup,
  title={Warmup Generations: A Task-Agnostic Approach for Guiding Sequence-to-Sequence Learning with Unsupervised Initial State Generation},
  author={Li, Senyu and Sun, Zipeng and Wang, Jiayi and Liu, Xue and Stenetorp, Pontus and Reddy, Siva and Adelani, David Ifeoluwa},
  booktitle={Proceedings of the 63rd Annual Meeting of the Association for Computational Linguistics (Volume 1: Long Papers)},
  pages={8870--8880},
  year={2025}
}

@inproceedings{akarajaradwong2025cold,
  title={Cold Starts and Hard Cases: A Two-Stage SFT-RLVR Approach for Legal Machine Translation (Just-NLP L-MT shared task)},
  author={Akarajaradwong, Pawitsapak and Chaksangchaichot, Chompakorn},
  booktitle={Proceedings of the 1st Workshop on NLP for Empowering Justice (JUST-NLP 2025)},
  pages={101--106},
  year={2025}
}

@article{wei2025advancing,
  title={Advancing multimodal reasoning via reinforcement learning with cold start},
  author={Wei, Lai and Li, Yuting and Zheng, Kaipeng and Wang, Chen and Wang, Yue and Kong, Linghe and Sun, Lichao and Huang, Weiran},
  journal={arXiv preprint arXiv:2505.22334},
  year={2025}
}

@inproceedings{chen2025optimizing,
  title={Optimizing Efficiency and Visual-Textual Alignment for LLM-Based Radiology Report Generation},
  author={Chen, Zailong and Gao, Peng and Lee, Yujian and Barthelemy, Johan and Zhou, Luping and Wang, Lei},
  booktitle={2025 IEEE International Conference on Multimedia and Expo (ICME)},
  pages={1--6},
  year={2025},
  organization={IEEE}
}

@inproceedings{gao2026remember,
  title={Remember Me: Bridging the Long-Range Gap in LVLMs with Three-Step Inference-Only Decay Resilience Strategies},
  author={Gao, Peng and Lee, Yujian and Zhang, Xiaofeng and Chen, Zailong and Zhang, Hui},
  booktitle={Proceedings of the AAAI Conference on Artificial Intelligence},
  volume={40},
  number={6},
  pages={4140--4148},
  year={2026}
}

@inproceedings{gao2025ncl,
  title={NCL-CIR: Noise-aware contrastive learning for composed image retrieval},
  author={Gao, Peng and Lee, Yujian and Chen, Zailong and Liu, Xubo and Zhang, Hui and Hu, Yiyang and Jing, Guquan},
  booktitle={ICASSP 2025-2025 IEEE International Conference on Acoustics, Speech and Signal Processing (ICASSP)},
  pages={1--5},
  year={2025},
  organization={IEEE}
}

@inproceedings{gao2025contextual,
  title={Contextual Reasoning for Robust Composed Image Retrieval with Vision-Language Models},
  author={Gao, Peng and Lee, Yujian and Liu, Xubo and Zhang, Hui and Chen, Zailong and Hu, Yiyang and Jing, Guquan and Lai, Yunting},
  booktitle={Proceedings of the 2025 International Conference on Multimedia Retrieval},
  pages={349--357},
  year={2025}
}




\end{document}